%% file: main.tex
\documentclass{article}

\usepackage{microtype}
\usepackage{graphicx}
\usepackage{subcaption}
\usepackage{booktabs} 

\usepackage{hyperref}

\usepackage[preprint]{icml2026}

\usepackage{amsmath}
\usepackage{amssymb}
\usepackage{mathtools}
\usepackage{amsthm}

\theoremstyle{plain}

\theoremstyle{definition}

\theoremstyle{remark}

\usepackage[textsize=tiny]{todonotes}

\usepackage{url}
\usepackage{multirow}
\usepackage[capitalise, nameinlink]{cleveref}
\usepackage{enumitem}
\usepackage{comment}
\usepackage[table]{xcolor}

\usepackage{makecell}
\usepackage{listings}
\usepackage{wrapfig}
\usepackage{pdfx}

\newcommand{\parta}{high-level policy}
\newcommand{\partb}{low-level policy}

\newcommand{\out}{control action}

\newcommand{\MethodName}[0]{SteerVLA}

\icmltitlerunning{SteerVLA: Steering Vision-Language-Action Models in Long-Tail Driving Scenarios}

\begin{document}

\twocolumn[
  \icmltitle{SteerVLA: Steering Vision-Language-Action Models in \\ Long-Tail Driving Scenarios}
  \icmlsetsymbol{equal}{*}

  \begin{icmlauthorlist}
    \icmlauthor{Tian Gao}{equal,stanford}
    \icmlauthor{Celine Tan}{equal,ucb}
    \icmlauthor{Catherine Glossop}{ucb}
    \icmlauthor{Timothy Gao}{ucb}
    \icmlauthor{Jiankai Sun}{stanford}
    \icmlauthor{Kyle Stachowicz}{ucb}
    \icmlauthor{Shirley Wu}{stanford}
    \icmlauthor{Oier Mees}{ucb,microsoft}
    \icmlauthor{Dorsa Sadigh}{stanford}
    \icmlauthor{Sergey Levine}{ucb}
    \icmlauthor{Chelsea Finn}{stanford}
  \end{icmlauthorlist}

  \icmlaffiliation{stanford}{Stanford University}
  \icmlaffiliation{ucb}{University of California, Berkeley}
  \icmlaffiliation{microsoft}{Microsoft}


  \icmlcorrespondingauthor{Tian Gao}{tiangao@stanford.edu}

  \icmlkeywords{Vision-Language-Action Models, Machine Learning, ICML}

  \vskip 0.3in
]
\printAffiliationsAndNotice{} 

\input{data/0_abstract}
\input{data/1_intro}
\input{data/2_related_work}

\input{data/3_method}
\input{data/4a_experimental_setup}

\input{data/4b_results}
\input{data/5_conclusion}
\input{data/6_ack}

\bibliography{bibtex_robotics}
\bibliographystyle{icml2026}

\input{data/7_app}

\end{document}

%% file: data/0_abstract.tex
\begin{abstract}
A fundamental challenge in autonomous driving is the integration of high-level, semantic reasoning for long-tail events with low-level, reactive control for robust driving. While large vision-language models (VLMs) trained on web-scale data offer powerful common-sense reasoning, they lack the grounded experience necessary for safe vehicle control. We posit that an effective autonomous agent should leverage the world knowledge of VLMs to guide a steerable driving policy toward robust control in driving scenarios.
To this end, we propose \MethodName{}, which leverages the reasoning capabilities of VLMs to produce fine-grained language instructions that steer a vision-language-action (VLA) driving policy.
Key to our method is this rich language interface between the high-level VLM and low-level VLA, which allows the high-level policy to more effectively ground its reasoning in the control outputs of the low-level policy. To provide fine-grained language supervision aligned with vehicle control, we leverage a VLM to augment existing driving data with detailed language annotations, which we find to be essential for effective reasoning and steerability. We evaluate \MethodName{} on a challenging closed-loop benchmark, where it outperforms state-of-the-art methods by \textbf{4.77} points in overall driving score and by \textbf{8.04} points on a long-tail subset. The project website is available at: \url{https://steervla.github.io/}.
\end{abstract}

\begin{figure*}
    \centering
    \includegraphics[width=\linewidth]{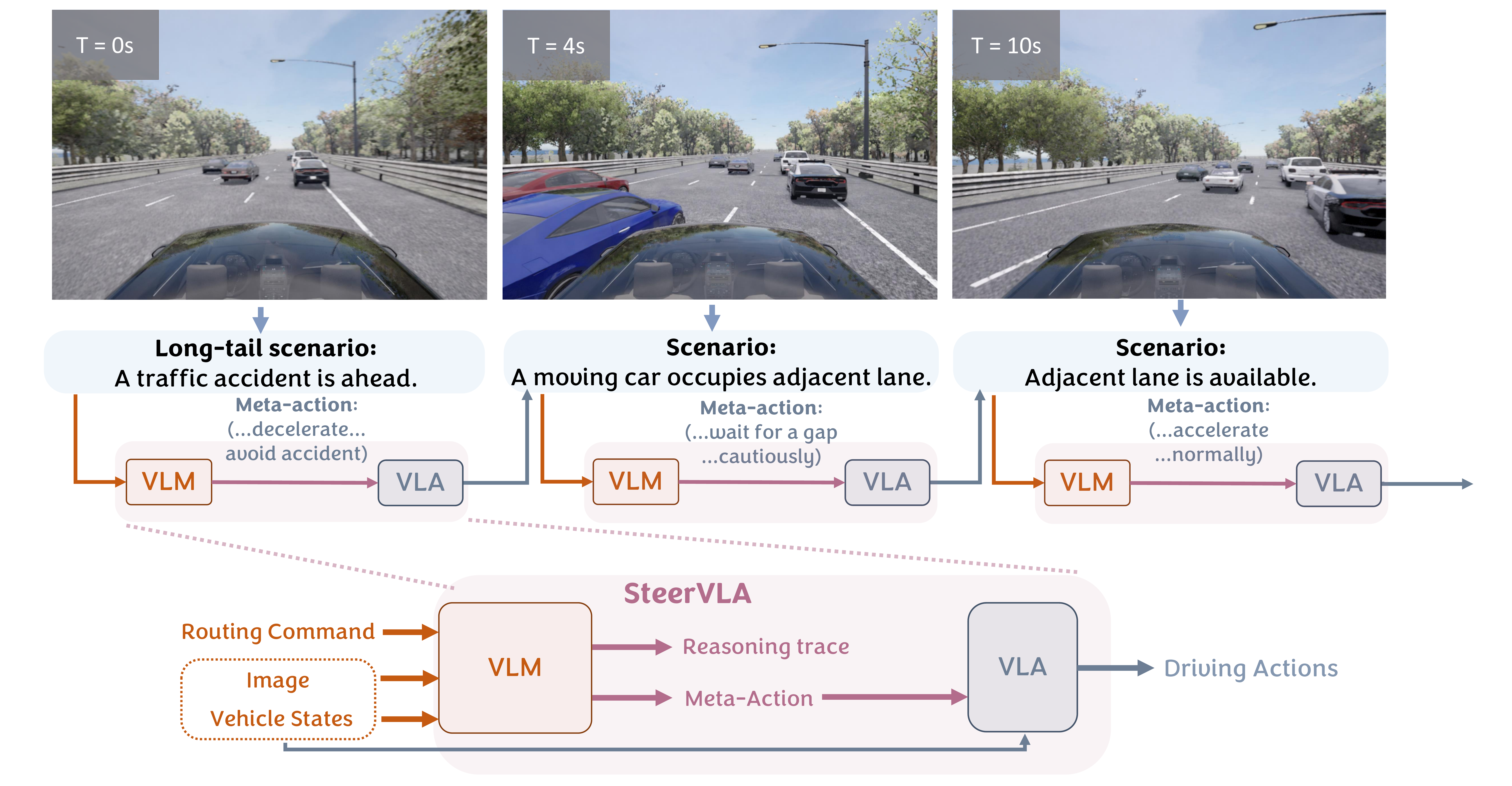}
    \vspace{-10pt}
    \caption{\textbf{\MethodName{} encountering long-tail scenarios}. \MethodName{} is able to quickly reason about and adapt to a traffic accident blocking the lane. It first slows down, then waits for a gap in the traffic, and merges when the lane is available.} 
    \label{fig:teaser}
    \vspace{-5pt} 
\end{figure*}

%% file: data/1_intro.tex
\section{Introduction}
\label{sec:intro}

Despite rapid progress in autonomous driving systems, long-tail scenarios remain particularly challenging due to their inherent scarcity in driving data and the complex reasoning they require. A truly autonomous vehicle must handle ambiguous traffic flow in construction zones, unpredictable pedestrian behavior, and blocked lanes due to accidents, as well as the compositions of these scenarios. For example, in~\cref{fig:teaser}, the vehicle encounters an accident blocking the lane. It must first reason about the scenario and recognize that it cannot simply continue in the same lane. It then needs to decide the best course of action while considering the other vehicles on the road. Handling these long-tail scenarios effectively is fundamentally important to building safe and robust driving systems~\citep{tokenizetheworld}.

Vision–language–action (VLA) models, derived from vision–language models (VLMs) and adapted to driving control via imitation learning, leverage strong semantic priors to generate embodied actions~\citep{brohan2023rt2visionlanguageactionmodelstransfer,kimOpenVLAOpenSourceVisionLanguageAction2024,zhou2025opendrivevla}. However, long-tail driving scenarios often require reasoning over rare events, implicit social norms, and broader common-sense knowledge that can be very hard to infer reliably from immediate visual driving cues alone. Pretrained VLMs encode such semantic knowledge, but effectively applying it in driving depends on how these inferences are grounded in control. We focus on enabling semantic reasoning from VLMs to guide driving behavior in VLA policies, allowing effective use of pretrained knowledge in complex and long-tail driving scenarios.

We present \MethodName{}, a novel framework for VLA-based driving policies effective in both normal and long-tail scenarios. Our key insight is to steer VLA control using VLM reasoning through:
1) a \parta{}, fine-tuned from a pretrained VLM, that performs semantic and common-sense reasoning to analyze driving scenarios based on camera images, routing commands (e.g., ``Turn left at the next intersection'') from navigation APIs, and historical vehicle states. This model outputs reasoning traces and \emph{meta-actions}—driving instructions for ego-vehicle motion (e.g., ``accelerate and make a wide left turn, cautiously monitoring the junction.''), and
2) a \partb{}, fine-tuned from a pretrained VLM to generate precise \out{}s conditioned on meta-actions.
This design leverages the powerful reasoning capabilities of VLMs while producing fine-grained outputs for driving control in VLAs.

The key challenges are enabling the high-level policy to reason over diverse driving scenarios and generate detailed commands, as well as training a low-level policy to reliably execute them in a steerable manner. For example, if an accident blocks the road and the car ahead enters the opposite driving lane, the \parta{} should reason that ``The lane appears shared bidirectionally since the vehicle in front moved into the oncoming lane. Proceed cautiously'' and output conservative acceleration from a stopped state. Given camera images and vehicle states, the \partb{} maps meta-actions to precise \out{}s such as speed control and steering angle. 

However, difficulty arises from the scarcity of natural language supervision in driving datasets, where language annotations are often missing or coarse, and rarely grounded in driving control.
To provide high-quality language supervision, we design an automatic data generation pipeline that constructs language supervision explicitly grounded in control. Given driving scenes and their associated driving trajectories, the pipeline guides a VLM to generate or refine meta-actions that describe the underlying driving behaviors over a temporal window. The pipeline also augments images with mid-level representations (bounding boxes or trajectory projections) to help the VLM better reason about spatial relationships and align its understanding with driving control.
Rather than relying on generic commands such as ``decelerate due to the stop sign'', we enrich meta-actions with details derived from trajectories, including motion intensity and directional adjustments. This results in grounded descriptions such as ``decelerate rapidly and cautiously make a slight right adjustment before stopping for a sign,'' which better steer the low-level policy. 
Using this automatically generated supervision, we fine-tune the high-level policy to reason over complex scenes and produce well-grounded meta-actions. We train the low-level policy to follow these detailed meta-actions and imitate the safe driving behavior in our training data. The high-level policy can then control the low-level policy at the meta-action level, allowing for safe, intelligent responses in driving scenarios that require complex reasoning.

This paper introduces a training and data generation framework that enables semantic reasoning to steer driving control, leading to strong performance in long-tail scenarios. We evaluate \MethodName{} on the Bench2Drive~\citep{jia2024bench2drivemultiabilitybenchmarkingclosedloop} benchmark in the CARLA~\citep{dosovitskiy2017carlaopenurbandriving} simulator. To test long-tail performance specifically, we identify 11 long-tail scenarios in Bench2Drive, which we term Bench2Drive-LongTail. We find that \MethodName{} outperforms state-of-the-art methods by \textbf{4.77} points in driving score on Bench2Drive overall, and notably by \textbf{8.04} points on Bench2Drive-LongTail, confirming that grounded reasoning improves generalization in long-tail scenarios. 

%% file: data/2_related_work.tex
\section{Related Work}
\label{sec:related_work}
We review related work in three areas: VLA–based driving models, methods that incorporate reasoning into driving models, and data labeling for driving data. 

\textbf{Vision-language-action models in driving.} Although autonomous driving has traditionally consisted of methods that use a stack of perception, prediction, and planning modules~\citep{hu2023planningorientedautonomousdriving, huang2021learning, sun2021neuro}, massive progress has been made with end-to-end imitation learning methods that directly map multi-modal inputs to driving commands~\citep{feng2025uniplan, Nguyen2025OpenXAV, zheng2025diffusionbasedplanningautonomousdriving, hegde2025distillingmultimodallargelanguage}. These methods generally excel in generic driving scenarios, but can struggle to generalize to long-tail scenarios, as these are not well-covered in driving data. 

Several works have gone beyond training end-to-end policies from scratch and leverage large language and vision-language models to leverage their pre-trained capabilities. Various works fine-tune pre-trained large language models on driving data~\citep{jia2023adriver,yuan2024rag,hwang2024emma,arai2025covla,zhou2025opendrivevla,fu2025orion,gao2025langcoop,zhou2025autovla, wang2023drivemlm,shao2024lmdrive}. Some~\citep{chen2024driving,xu2024drivegpt4interpretableendtoendautonomous} integrate multimodal inputs, such as images, by projecting them into token space, while others~\citep{mao2023gpt, mao2023language, qian2025agentthink} adapt pre-trained VLMs as motion planners through text-based fine-tuning. Inspired by the success of pretrained vision-language models (VLMs), several works have introduced \textit{vision-language-action} (VLA) models \citep{brohan2023rt2visionlanguageactionmodelstransfer}, which consist of a VLM backbone fine-tuned to produce robot actions~\citep{kimOpenVLAOpenSourceVisionLanguageAction2024}. These models benefit from excellent cross-modal grounding between language and vision, enabling the transfer of internet-scale semantic knowledge from their pre-training data. However, a key challenge for these methods is retaining the strong capabilities learned during pre-training, which can be lost when transferring to the domain of action prediction, a task very different from those found in VLM pre-training~\citep{driess2025knowledgeinsulatingvisionlanguageactionmodels}. While some of these methods fine-tune VLMs with an action head~\citep{hwangEMMAEndtoEndMultimodal2024, zhouOpenDriveVLAEndtoendAutonomous2025a,tokenizetheworld, renz2025simlingovisiononlyclosedloopautonomous} to mitigate this issue, we explicitly use a hierarchical model, allowing the high-level policy training to stay closer to VLM pre-training tasks. Moreover, we develop an auto-labeling pipeline for autonomous driving data, real or simulated, to provide dense language labels in the form of reasoning traces and detailed meta-action labels, to train the high-level and low-level policies. We find that while the hierarchical structure is essential for retention of the reasoning capabilities of the base VLM in the high-level policy, these dense labels are what allow for effective communication between the policies, which is key to \MethodName{}'s performance in long-tail scenarios.

\textbf{Reasoning in Autonomous Driving.} 
Recent works have sought to imbue VLAs with reasoning capabilities~\citep{Zawalski24-ecot, zhao2025cotvlavisualchainofthoughtreasoning, mu2023embodiedgptvisionlanguagepretrainingembodied, shi2024yell, belkhale2024rthactionhierarchiesusing, chen25training, argos25, liu2026unifiedembodiedvlmreasoning, ye2025vlar1enhancingreasoningvisionlanguageaction} to improve generalization and compositional task-following.
In the driving domain, reasoning has been primarily used in the form of chain-of-thought steps~\citep{zhou2025autovla,qian2025agentthink,wang2025cot4advisionlanguageactionmodelexplicit,luo2025adathinkdriveadaptivethinkingreinforcement, hegde2025distillingmultimodallargelanguage,xu2024drivegpt4interpretableendtoendautonomous,renz2025simlingovisiononlyclosedloopautonomous}, casting reasoning as detecting other vehicles, describing the scene, performing question-answering tasks, or providing explainability or justification signals. 
While these methods improve reasoning or generalization, they remain largely descriptive. In contrast, we use our auto-labeling pipeline to generate both descriptive reasoning traces, specifically including the states of other vehicles and traffic sign information, and detailed \emph{prescriptive} meta-action labels.
Most similar to our work is SimLingo~\citep{renz2025simlingovisiononlyclosedloopautonomous}, which also focuses on long-tail driving scenario capabilities and achieves state-of-the-art performance on the Bench2Drive benchmark. However, SimLingo relies on access to ``action dreaming'' data. This consists of safe and unsafe trajectories collected in CARLA with access to privileged information, exposing the policy to counterfactual scenarios beyond expert demonstrations. \MethodName{}'s auto-labeling pipeline does not require access to privileged simulation information, and can be easily transferred to real-world data. We find that we can achieve improved long-tail performance with the labels generated by our auto-labeling pipeline in combination with a hierarchical policy structure. We also do not rely on additional data to improve reasoning and steerability, but achieve this through our hierarchical architecture and detailed meta-action labels to steer the low-level policy, especially in scenarios where reasoning about other agents is required, such as the ``blocked intersection'' and ``construction zone'' scenarios shown in~\cref{fig:longtail_bench2drive} and~\cref{fig:case_study}.

\textbf{Data labeling for Autonomous Driving.}
Several works have aimed to label driving data with language to improve interpretability and reasoning capabilities in driving models. Some use a purely manual labeling process~\citep{xu2020explainable, Deruyttere_2019, malla2022dramajointrisklocalization, wu2025languagepromptautonomousdriving}, which introduces high overhead, but can result in more natural and realistic labels. Other works use a mixture of VLM-generated labels and human verification to create a more scalable labeling pipeline~\citep{simaDriveLMDrivingGraph2024, inoue2023nuscenesmqaintegratedevaluationcaptions, he2025carscenessemanticvlmdataset}. \MethodName{} is trained on data generated from our fully automatic labeling pipeline and does not require human supervision. Many of these works focus on visual question-answering tasks~\citep{simaDriveLMDrivingGraph2024, inoue2023nuscenesmqaintegratedevaluationcaptions} or explanations of driving behavior~\citep{xu2020explainable, malla2022dramajointrisklocalization}, aiming to improve scene understanding and interpretability but with limited focus on precisely prescribing the actions the vehicle should take. \MethodName{}'s auto-labeling pipeline aims to extract information from driving data and organize it into comprehensive reasoning traces and detailed meta-action labels. 
These labels go beyond the typical commands used in prior work, such as ``Accelerate'' or ``Turn left''. Instead, we augment these commands with the manner in which these behaviors should be executed, for example, ``Accelerate cautiously with a slight left adjustment to avoid the construction site''. To further enhance the VLM's spatial reasoning during labeling, we overlay mid-level representations (bounding boxes or trajectory projections) onto the driving images. By training the high-level policy to generate these information-rich meta-actions and the low-level policy to follow them, we can effectively ground the high-level reasoning into low-level control.

%% file: data/3_method.tex
\section{Preliminaries}

\begin{figure*}[t]
    \centering
    \includegraphics[width=\linewidth]{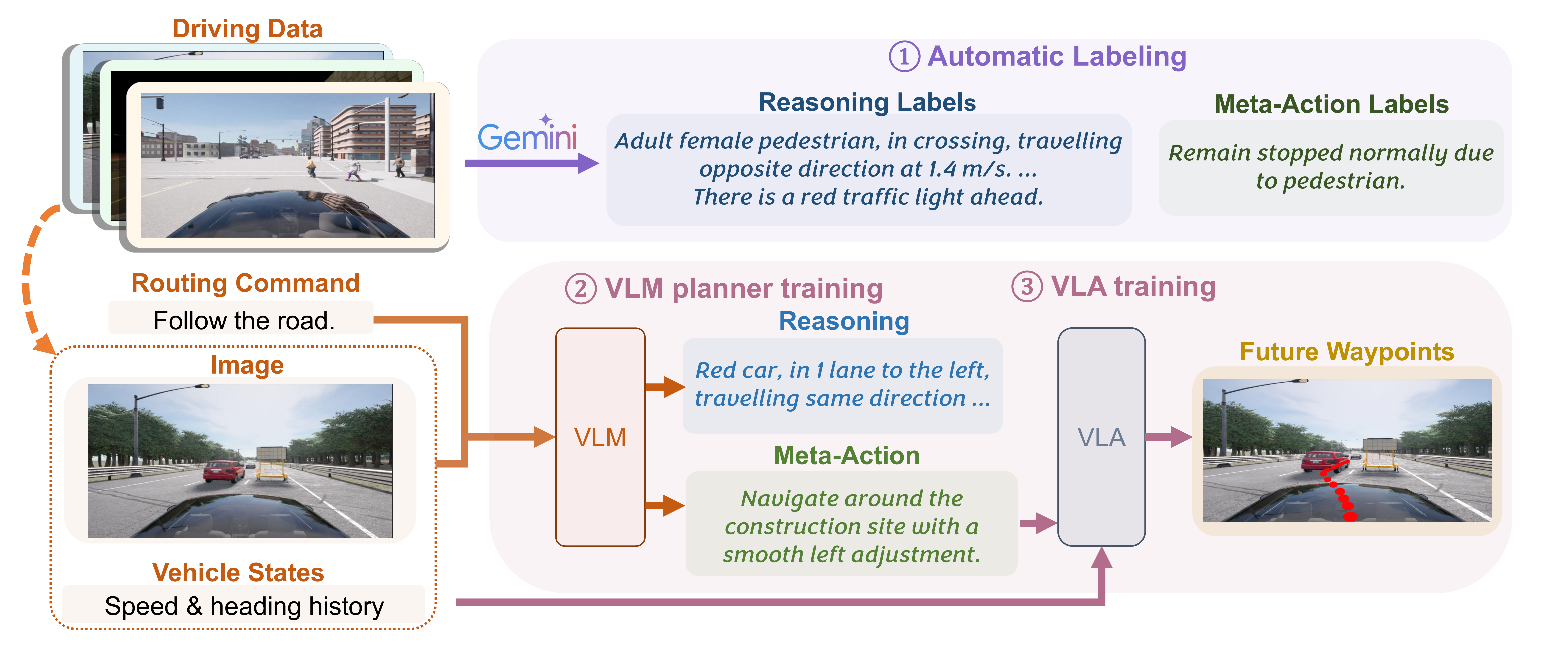}
    \caption{\textbf{Framework overview}. Our model reasons over the driving context to produce reasoning traces and fine-grained meta-actions that guide driving control. A VLM generates structured semantic guidance, which steers a VLA policy in predicting future waypoints. To supervise reasoning and meta-action generation, we introduce an automatic data labeling pipeline that derives fine-grained language supervision from driving trajectories, improving alignment between language and control.}
    \label{fig:method}
    \vspace{-5pt} 
\end{figure*}

\textbf{Problem statement.} We formulate autonomous driving as a sequential decision-making problem. We assume access to a standard navigation system that provides high-level routing commands to guide the vehicle toward a destination. At each timestep $t$, the agent receives an observation $o_t = \{ I_t, q_{t-k+1:t}\}$ and a routing command $\ell_t$, where $I_t$ is the current front-view camera image, $q_{t-k+1:t}$ denotes the recent history of ego vehicle states (e.g., past speeds and headings over the last $k$ steps), and $\ell_t$ is provided by a navigation system (e.g., turn-by-turn guidance such as ``turn left in 50\,m''). The objective is to predict a chunk of future actions $A_t = [a_t, a_{t+1}, \ldots, a_{t+H-1}]$ that specify low-level control signals (e.g., future waypoints) over a horizon $H$~\citep{zhao2023learning}. A driving policy is therefore a conditional distribution $\pi(A_t \mid o_t, \ell_t)$ that maps the current observation and routing command to a distribution over action chunks. Training proceeds by maximizing the likelihood of expert demonstrations: $\max_{\theta}\ \mathbb{E}_{(A_t, o_t, \ell_t)\sim D}\,[\log \pi_{\theta}(A_t \mid o_t, \ell_t)]$, where $D$ is a dataset of expert driving trajectories paired with synchronized routing commands.

\textbf{Driving vision-language-action models.} Recent driving VLA models~\citep{hwangEMMAEndtoEndMultimodal2024, zhouOpenDriveVLAEndtoendAutonomous2025a} learn a direct mapping from routing commands $\ell_t$ and visual observations $I_t$ to driving control actions $A_t$, typically by fine-tuning a pretrained vision–language model on driving data. Some approaches, such as SimLingo~\citep{renz2025simlingovisiononlyclosedloopautonomous}, generate a language-based description of the intended driving behavior (i.e., a meta-action) as a chain-of-thought step before producing driving actions. Our low-level policy in CARLA builds upon SimLingo~\citep{renz2025simlingovisiononlyclosedloopautonomous}, which uses InternVL2-1B~\citep{chen2024internvl} as the pretrained VLM backbone and represents actions as future waypoints. Waypoint prediction is performed by lightweight MLP heads on top of the VLM outputs and is trained using a Smooth L1 loss.

\section{\MethodName}
\label{sec:method}
In this section, we introduce \MethodName{}, a framework that leverages VLM semantic reasoning to steer a VLA policy toward grounded and context-aware driving control. An overview of \MethodName{} is shown in~\cref{fig:method}.

\subsection{Steering VLA Control using VLM Reasoning}
We focus on the challenge of long-tail driving scenarios, where rare and unanticipated events require strong generalization and common-sense reasoning from the policy. VLAs are a strong backbone for driving because they combine semantic grounding from vision–language pretraining with domain-specific adaptation obtained via imitation learning on driving data. Building on this capability, we leverage the reasoning and semantic inference abilities of VLMs and ground these inferences in driving control through fine-grained meta-actions that steer a VLA policy. Concretely, a high-level policy first reasons about the driving scene, historical vehicle states, and routing command to produce a meta-action $m_t$, accompanied by a short reasoning trace $c_t$ that reasons over driving scenes and helps the policy generate more appropriate meta-actions. Formally, given observation $o_t = \{I_t, q_{t-k+1:t}\}$ with image $I_t$, historical ego vehicle states $q_{t-k+1:t}$, and routing command $\ell_t$, the high-level policy outputs $(c_t, m_t) \sim \pi_{hl}(c_t, m_t \mid o_t, \ell_t)$. The low-level VLA then predicts future waypoints $A_t = [a_t, a_{t+1}, \ldots, a_{t+H-1}]$ conditioned on both the observation and the meta-action, i.e., $A_{t} \sim \pi_{ll}(A_{t} \mid o_t, m_t)$. This design improves generalization by offloading high-level reasoning to the high-level policy, while allowing the VLA to specialize in fast and accurate waypoint prediction conditioned on the high-level’s instructions. 

\textbf{High-level policy.} We finetune the high-level policy $\pi_{hl}(c_t, m_t|o_t, \ell_t)$ with a pre-trained VLM as the base model. Our dataset generation pipeline, which we introduce in ~\cref{sec:auto-label}, provides supervision for the high-level policy outputs.  The VLM’s strong semantic priors allow the high-level policy to reason about the vehicle's surroundings and encode rich contextual information into its predicted meta-actions, enabling more context-aware action predictions. We structure the query to the VLM as a visual question-answering problem by providing the current visual observation $I_t$, a six-second-long history of ego states $q_{t-k+1:t}$ (speed and heading) sampled at 0.5 Hz where $k=3$, and a routing command $\ell_t$. We train the model via a next-token prediction objective to generate a chain-of-thought reasoning trace $c_t$ describing the positions and movement of critical agents in the scene, followed by an appropriate meta-action $m_t$.

\textbf{Low-level VLA policy.} Instead of using a general routing command as language input, our low-level policy $\pi_{ll}(A_t|o_t,m_t)$ is steered by fine-grained meta-actions $m_t$ generated by the high-level policy $\pi_{hl}(c_t, m_t|o_t, \ell_t)$ (see \cref{sec:auto-label} for details on generating meta-action labels). We do not provide the reasoning trace as input to the low-level policy. The high-level policy already uses the reasoning trace as chain-of-thought to synthesize a well-grounded meta-action that distills the contextual and semantic information required for control. Conditioning the low-level policy only on the meta-action allows it to focus on precise execution and instruction following rather than semantic reasoning.

\subsection{Generating language labels for \MethodName}
\label{sec:auto-label} 
To address the scarcity of fine-grained, control-grounded language supervision in driving datasets, we develop a fully autonomous labeling pipeline. We perform a two-stage query to Gemini 2.5 Flash-Lite \citep{geminiteam2025geminifamilyhighlycapable} that first identifies the baseline action taken by the vehicle (e.g., changing lanes, continuing straight), then uses trajectory information to enrich the language labels with fine-grained behavioral details explicitly tied to driving control. First, in our baseline categorization query, we provide Gemini 2.5 Flash-Lite with a grounded representation of the vehicle's action using a projection of the vehicle's future trajectory over a front-facing camera view. We then perform a refinement step by providing the VLM with the vehicle's speed and course over time to produce a nuanced description of the vehicle's action. For example, we transform the original label ``the car is continuing straight'' into the more fine-grained ``the car normally accelerates, then maintains speed while subtly drifting right''. This refinement step is crucial for passing as much information as possible to the low-level policy and can be applied to any existing language-labeled driving dataset, allowing us to augment these data with additional information that can improve steerability and performance of the driving policy. 

To improve the reasoning capabilities of the high-level policy, we additionally generate reasoning trace labels for each trajectory in the training data. These traces describe the scene and characterize the motion of other agents, serving as chain-of-thought supervision that guides the prediction of meta-actions. More details on the prompts used for our auto-labeling pipeline are provided in~\cref{app:label_refinement} of the appendix.  

\subsection{Implementation Details}
For closed-loop evaluation in the CARLA simulator, we follow the recipe from SimLingo~\citep{renz2025simlingovisiononlyclosedloopautonomous} to train the driving policy, building both the high-level and low-level policies upon InternVL2-1B~\citep{chen2024internvl} as the pretrained VLM. Future actions are represented by two types of waypoints: (1) time-based waypoints sampled at 4 Hz, which determine target speed and temporal motion, and (2) geometry-based waypoints sampled at 1 m intervals, which describe the planned path and guide steering. These waypoints are converted into driving controls via PID controllers.

We apply the refinement and reasoning trace generation processes described in~\cref{sec:auto-label} to the SimLingo dataset. To construct reasoning trace labels for the high-level policy, we leverage available 3D bounding box annotations to identify agents within the ego vehicle's field of view. We then transform relevant attributes, such as velocity and position relative to the ego vehicle, into structured natural language descriptions (e.g., ``Red car, in one lane to the left, traveling same direction, at 6.1 m/s.'').

We synchronize the high- and low-level policies by querying them at the same frequency. At each step, the low-level policy waits for the high-level to generate a meta-action, which it then uses as language input to produce driving actions. Specifically, we query the framework at 20 Hz in the CARLA simulator.

%% file: data/4a_experimental_setup.tex
\section{Experimental Results}
\label{sec:exps}

\begin{figure*}
    \includegraphics[width=\textwidth]{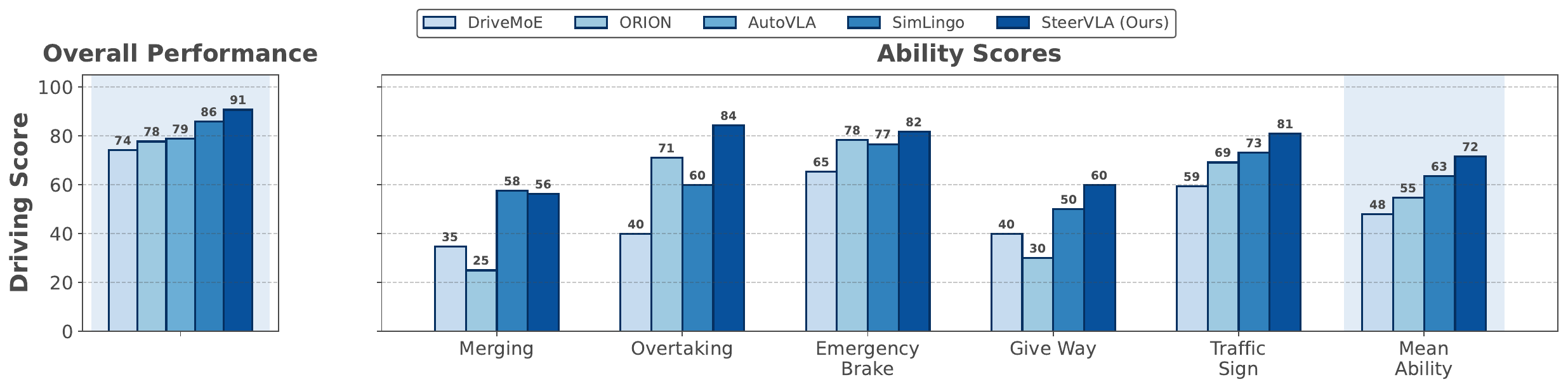}
\caption{\textbf{Closed-loop evaluation of \MethodName{} on Bench2Drive.} On Bench2Drive, we report overall performance and per-ability scores for \MethodName{} across five advanced urban driving skills. \MethodName{} significantly outperforms prior approaches, benefiting from improved reasoning and instruction-following capabilities.}
\label{fig:bench2drive}
\end{figure*}

\begin{figure*}
    \includegraphics[width=\textwidth]{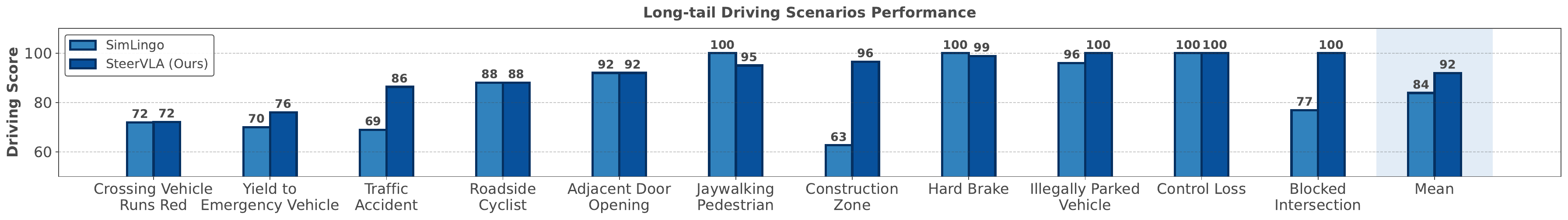}
\caption{\textbf{Closed-loop evaluation of \MethodName{} on Bench2Drive-LongTail.} We compare \MethodName{} with the state-of-the-art method SimLingo on Bench2Drive-LongTail. \MethodName{} exhibits larger performance gains in long-tail scenarios, likely because these cases require more complex reasoning and more precise control.}
\vspace{-5pt}
\label{fig:longtail_bench2drive}
\end{figure*}

Our experiments are designed to address the following key research questions:

\vspace{-0.1cm}
\begin{itemize}[leftmargin=30pt]
    \setlength{\parskip}{0cm}
    \setlength{\itemsep}{0cm}
    \item[{\bf RQ1:}]  How does \MethodName{} perform in simulated closed-loop driving under diverse traffic conditions, particularly in long-tail scenarios?
    \item[{\bf RQ2:}]  How effectively does \MethodName{} reason about complex scenes and follow driving instructions (i.e., meta-actions)?
    \item[{\bf RQ3:}]  How effective is each component in \MethodName{}?
    \item[{\bf RQ4:}]  Can \MethodName{} generalize to real-world driving data? 
\end{itemize}

\subsection{Experimental Setup}
The majority of our experiments use the CARLA simulator to perform large-scale closed-loop evaluation of \MethodName{} under diverse driving conditions. We evaluate \MethodName{} on the Bench2Drive~\citep{jia2024bench2drivemultiabilitybenchmarkingclosedloop} benchmark, which contains 220 driving scenarios in 12 towns, including adverse weather and lighting conditions, such as fog, nighttime driving, and various long-tail driving scenarios, such as construction sites, traffic accidents ahead, and jaywalking pedestrians. We also construct a benchmark Bench2Drive-LongTail consisting of a long-tail scenario subset from Bench2Drive, described in~\cref{sec:closed_loop} with additional detail in~\cref{app:long-tail}, to further study the performance of \MethodName{} on long-tail scenarios.

\paragraph{Training dataset.} We use the driving trajectories and meta-action labels from the SimLingo~\citep{renz2025simlingovisiononlyclosedloopautonomous} dataset, and further apply our data augmentation pipeline to generate reasoning traces and refine the meta-actions into more fine-grained descriptions. We train both the high- and low-level policies on this dataset.

\paragraph{Policy deployment.} We run \MethodName{} at 20 Hz in the CARLA simulator. On a single NVIDIA L40 GPU, \MethodName{} incurs an inference latency of 2.51 s. As our closed-loop evaluation is conducted entirely in simulation, we do not optimize inference efficiency in this work. For real-world deployment, we plan to reduce inference latency using standard acceleration techniques, such as KV caching, in future work.

\paragraph{Evaluation metrics.}
We evaluate closed-loop performance using the driving score, following the official CARLA metric. Driving score jointly measures task completion and safety by combining route completion with penalties for traffic infractions. Specifically, for each route, the route completion percentage is multiplied by penalties corresponding to the severity of infractions incurred. This metric captures both driving progress and robustness to safety violations.

\paragraph{Baselines.}
We evaluate several recent vision-language-action (VLA) baselines. Full descriptions of these methods are provided in~\cref{app:baselines}. We compare \MethodName{} to SimLingo~\citep{renz2025simlingovisiononlyclosedloopautonomous}, the current top method on the CARLA 2.0 Leaderboard trained with counterfactual data to improve language following, DriveMoE~\citep{yang2025drivemoemixtureofexpertsvisionlanguageactionmodel}, a mixture-of-experts method, an alternative to the hierarchical structure we introduce, ORION~\citep{fu2025orion}, an end-to-end method that focuses on long-term history aggregation and improves driving reasoning with question-answering as a co-training task rather than keeping a distinct high-level policy to maintain reasoning capabilities, and AutoVLA~\citep{zhou2025autovla}, which uses a pretrained VLM with a physical action codebook. These methods present alternative methods to retain reasoning capabilities or improve reasoning for driving tasks from a pretrained VLM.

%% file: data/4b_results.tex
\begin{table}[]
\small
\centering
\renewcommand{\arraystretch}{1.2}
\setlength{\tabcolsep}{5pt}

\begin{tabular}{lcc|c}
\toprule
\textbf{Method} & \textbf{\makecell[c]{Architecture}} & \textbf{Language Labels} & \textbf{\makecell[c]{Driving \\ Score}} $\uparrow$ \\
\midrule
SimLingo & VLA & Meta-actions & 85.94 \\
\MethodName{}  & VLM-VLA & Meta-actions & 88.81 \\
\midrule
\rowcolor{blue!10}
\MethodName{} & VLM-VLA & \makecell[c]{Refined meta-actions \\ + Reasoning traces} & \textbf{90.71} \\
\bottomrule
\end{tabular}

\caption{\textbf{Ablation study of \MethodName{} components on Bench2Drive.} Our results show that \MethodName{} benefits substantially from grounded semantic reasoning and fine-grained meta-actions produced by a high-level policy that effectively steers low-level control. This is enabled by our data labeling pipeline, which aligns fine-grained meta-action and reasoning supervision with low-level control signals extracted from driving trajectories.}
\vspace{-8mm}
\label{tab:design_decisions}
\end{table}

\begin{figure*}[htbp]
    \centering
    \begin{subfigure}{1.0\textwidth}
        \includegraphics[width=\linewidth]{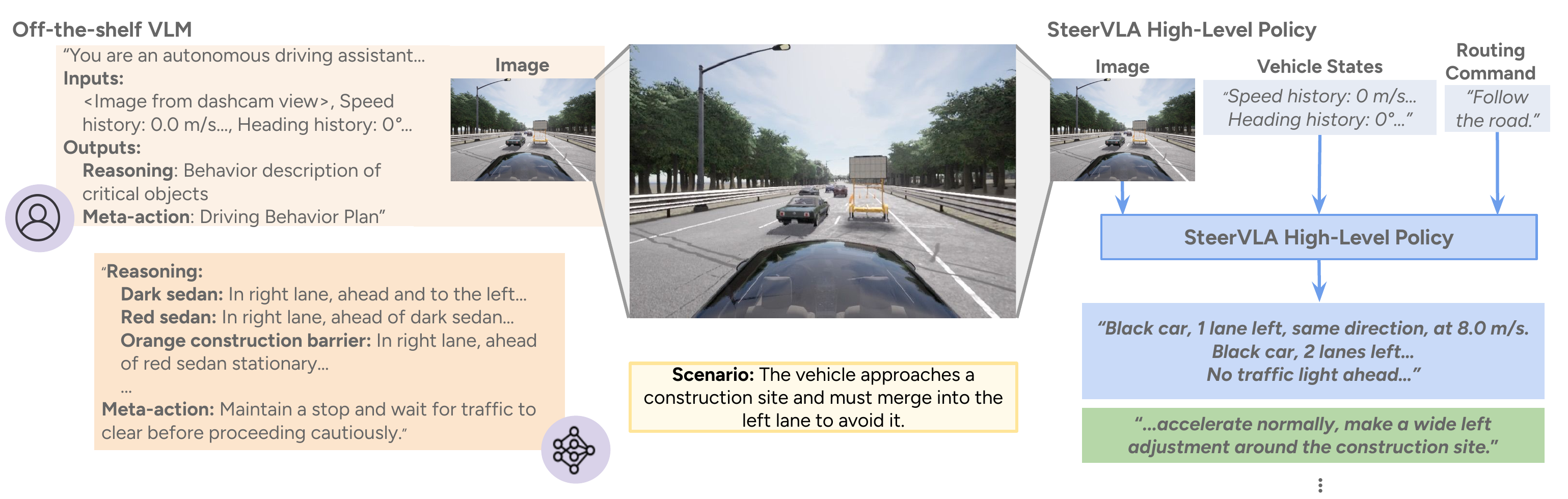}
        \caption{\textbf{Comparison of \MethodName{} with off-the-shelf VLM in reasoning capability.} We compare the reasoning capability in \MethodName{} high-level policy with Gemini 2.5 Flash Lite (prompting details in ~\cref{appendix:zero_shot_gemini}). When faced with a long-tail scenario, the off-the-shelf VLM (left) can roughly reason about the state of the scene but struggles to reason about the immediate actions the ego-vehicle should take under current conditions. Conversely, \MethodName{} (right) produces both descriptive reasoning and can aptly generate a meta-action to navigate around the construction site.}
        \label{fig:case_study_reasoning}
    \end{subfigure}
    \\
    \begin{subfigure}{\textwidth}
        \centering
        \includegraphics[width=1.0\linewidth]{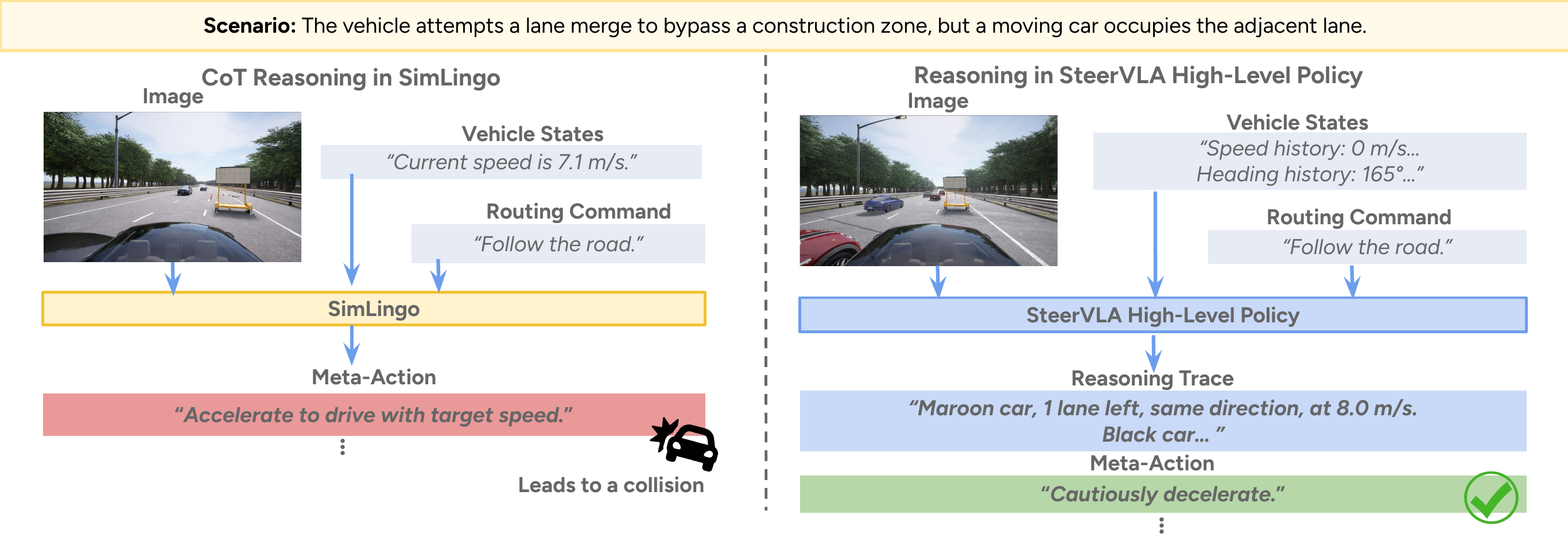}
        \caption{\textbf{Comparison of \MethodName{} with SimLingo in reasoning capability.} The detailed reasoning traces and meta-actions used to train \MethodName{} enable flexible environmental inference and timely decision-making when a nearby vehicle does not decelerate to give way during a lane change, whereas SimLingo fails to generate timely meta-actions and collides with another vehicle.}
        \label{fig:case_study_simlingo_reasoning}
    \end{subfigure}
    \\
    \begin{subfigure}{\textwidth}
        \centering
        \includegraphics[width=1.0\linewidth]{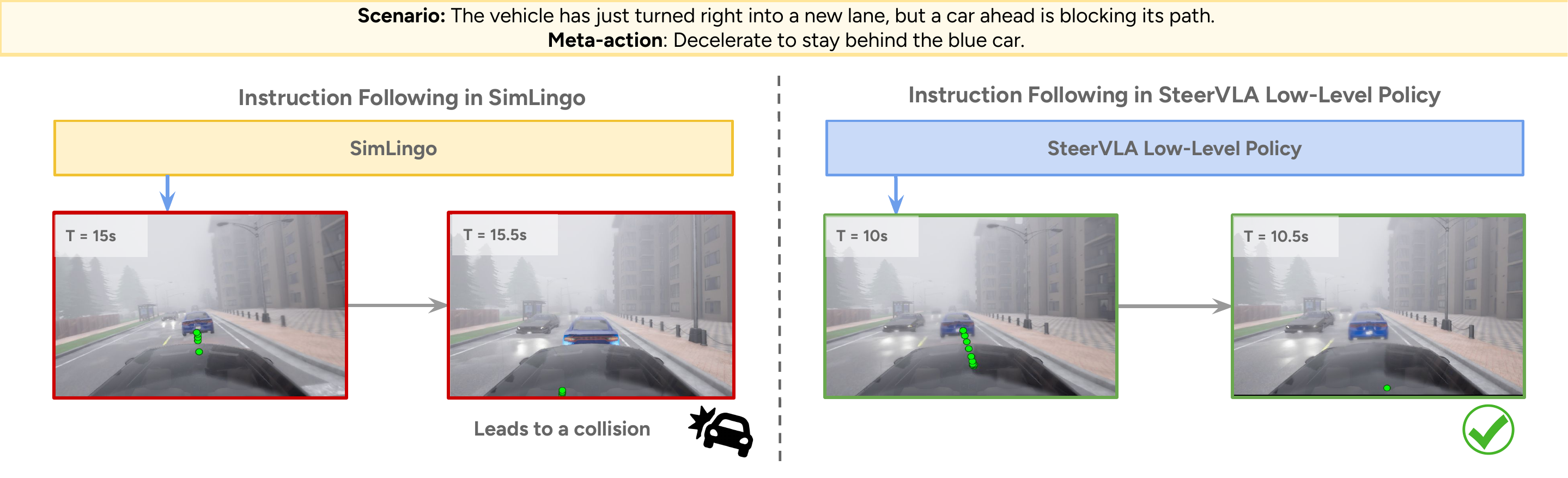}
        \caption{\textbf{Comparison of \MethodName{} with SimLingo in instruction following capability.} After turning right into a new lane where the vehicle ahead blocks the path, both methods output the meta-action decelerate to follow the blue vehicle. \MethodName{} decelerates immediately, while SimLingo’s delayed deceleration leads to a collision. Green points show predicted future waypoints sampled at 4 Hz.}
        \label{fig:case_study_simlingo_instruction}
    \end{subfigure}
    \vspace{-10pt}
    \caption{\textbf{Long-tail scenario case study.} We analyze how \MethodName{} reasons and acts in long-tail driving scenarios. In \textbf{(a)}, we compare \MethodName{} with an off-the-shelf VLM, showing that \MethodName{} produces both descriptive reasoning and actionable meta-actions. In \textbf{(b)}, we compare \MethodName{} with SimLingo in a lane-change interaction where an adjacent vehicle does not give way, highlighting \MethodName{}’s ability to make timely high-level decisions. In \textbf{(c)}, we evaluate instruction-following behavior when the lane ahead is blocked, where \MethodName{} executes the deceleration meta-action immediately while SimLingo exhibits delayed control.}
    \label{fig:case_study}
\end{figure*}

\subsection{Evaluating \MethodName{} on Driving Performance}
\label{sec:closed_loop}
Towards answering \textbf{Q1}, we evaluate \MethodName{} closed-loop on the Bench2Drive benchmark. We additionally extract a subset of routes from Bench2Drive to observe the long-tail reasoning capabilities of \MethodName{} and its ability to act appropriately in these scenarios. Results are shown in~\cref{fig:bench2drive} and in~\cref{fig:longtail_bench2drive} with a detailed table of results in~\cref{app:raw_values}. We also include discussion on failure cases in~\cref{app:failure_cases}.

\textbf{Driving performance on Bench2Drive.} \cref{fig:bench2drive} demonstrates that \MethodName{} has strong performance on the Bench2Drive benchmark, achieving a better driving score than the next best baseline, SimLingo, by 4.77, and outperforms it in all ability categories, except merging. We observe that \MethodName{} tends to outperform SimLingo in highly dynamic scenarios (i.e., merging into the oncoming lane around a construction zone, or navigating an accident on the road). \MethodName{}'s reasoning trace structure guides the policy to make conjectures about the movement intent of the other agents within the scene, enabling \MethodName{} to prepare and preemptively react to adversarial behavior. This is also made apparent through the case study presented in~\cref{fig:case_study}.

\textbf{Driving performance in long-tail scenarios.} On Bench2Drive-LongTail, \MethodName{} shows clear advantages over the strongest baseline, SimLingo. \cref{fig:longtail_bench2drive} demonstrates that \MethodName{}, on average, outperforms SimLingo by 8.04 in driving score, with especially large deltas in the construction zone, traffic accident, and blocked intersection scenarios. Long-tail scenarios demand rich semantic reasoning and precise control. In these scenes, inferences about out-of-distribution objects and the behavior of other agents must be carefully interpreted to navigate the changes in traffic flow from typical scenarios. \MethodName{} is able to address this by explicitly reasoning about the states of other agents and traffic signs, resulting in detailed meta-actions that can guide VLA to cautiously but decisively act when it is safe to do so.

\subsection{Case Study in Long-Tail Scenarios}
Towards answering \textbf{Q2}, we perform case studies of \MethodName{} on challenging long-tail scenarios to observe its reasoning and instruction following capabilities, shown in~\cref{fig:case_study}. We first compare \MethodName{}'s high-level policy reasoning to Gemini 2.5 Flash Lite, the same VLM we use in our autolabeling pipeline. In~\cref{fig:case_study_reasoning}, when faced with a construction site blocking a lane, while the off-the-shelf VLM can roughly reason about the scene state, it struggles to reason about the immediate actions the ego-vehicle should take. In contrast, \MethodName{} produces both descriptive reasoning and aptly generates a meta-action to navigate around the construction site. Comparing to SimLingo in~\cref{fig:case_study_simlingo_reasoning}, we observe that the detailed reasoning traces and meta-actions used to train \MethodName{} enable more dynamic, information-rich reasoning and timely decision-making. When a nearby vehicle does not decelerate during a lane change, \MethodName{} generates appropriate meta-actions by reasoning about other agents and traffic conditions, while SimLingo fails to react in time and collides. Furthermore, in~\cref{fig:case_study_simlingo_instruction}, when both methods correctly output a deceleration meta-action to follow a blocking vehicle, \MethodName{} executes immediately while SimLingo's delayed response leads to a collision. Overall, these case studies demonstrate that the reasoning traces and meta-actions of \MethodName{} enable state-of-the-art long-tail scenario performance through superior environmental reasoning and precise action timing.

\subsection{Ablations of \MethodName}
Towards answering \textbf{Q3}, we study the effectiveness of two key design choices in \MethodName{} on Bench2Drive: the VLM-VLA style architecture and the reasoning components (i.e. fine-grained meta-actions and reasoning traces).
We evaluate the effectiveness of a hierarchical architecture by comparing SteerVLA with SimLingo under the same meta-action supervision from the SimLingo dataset.
We assess the impact of fine-grained meta-actions and reasoning traces by comparing \MethodName{} trained with refined language labels generated by our data labeling pipeline against training with the original meta-action labels from SimLingo.
The results in \cref{tab:design_decisions} show that \MethodName{} benefits substantially from grounded semantic reasoning and fine-grained meta-actions produced by the high-level policy that effectively steer low-level control. This is enabled by our data labeling pipeline, which aligns fine-grained meta-action and reasoning supervision with low-level control signals extracted from driving trajectories.

\subsection{Open-Loop Evaluation on Real-World Data} 
\label{sec:open_loop}
Towards answering Q4, we additionally evaluate \MethodName{} in an open-loop setting on the NuScenes benchmark~\citep{nuscenes}, using stronger VLM backbones—Gemma3-4B~\citep{gemmateam2025gemma3technicalreport} as the high-level policy and PaliGemma-3B~\cite{beyer2024paligemmaversatile3bvlm} as the low-level policy. We adopt more powerful pretrained VLMs for this evaluation, as real-world driving scenarios require stronger visual semantic generalization and benefit from richer prior knowledge learned from large-scale real-image data. Besides, since NuScenes does not provide language annotations, we apply our auto-labeling pipeline to generate grounded meta-actions and reasoning traces directly from raw driving data. We emphasize that our primary evaluation focuses on closed-loop results on Bench2Drive, which provide a more meaningful assessment by capturing control dynamics, recovery behavior, and long-horizon interactions with the environment, whereas open-loop evaluation on NuScenes measures trajectory imitation on fixed data. \MethodName{} achieves comparable performance to existing methods on the NuScenes benchmark. Detailed experimental settings and results on NuScenes are provided in~\cref{app:nuscenes}.

%% file: data/5_conclusion.tex
\section{Discussion}
\label{sec:conclusion}
We presented \MethodName{}, a hierarchical VLA model for autonomous driving. Our approach decomposes driving into a high-level, language-based reasoning step and a low-level action generation step, with detailed meta-actions serving as the interface between the two. These meta-actions are generated from driving data through a fully automatic labeling pipeline. As a result, \MethodName{} can reason over complex driving scenarios and produce precise control outputs, achieving state-of-the-art performance in both general and long-tail driving settings.

While \MethodName{} demonstrates improved reasoning and steerability, it has several limitations that point to important future directions. The quality of our auto-labeling pipeline is constrained by the capabilities of the underlying VLM, particularly for temporally grounded understanding in video contexts. Our current system also uses only a single camera view, limiting scene coverage; extending to multi-view camera inputs would enhance spatial awareness and better match real-world autonomous vehicle sensor configurations.

Overall, we hope that our work represents a step toward real-world systems that can use common sense to deal with complex and unfamiliar situations. We expect that the capabilities of VLMs and other foundation models will continue to improve, providing better multi-modal reasoning in diverse scenarios, and grounding these capabilities in real-world actions would allow for increasingly robust autonomous systems. \MethodName{} represents a step toward this future.

%% file: data/6_ack.tex
\section*{Acknowledgements}
We thank Yuejiang Liu, Anikait Singh, William Chen, Priya Sundaresan, Joey Hejna, and others in the Stanford IRIS Lab, the Berkeley RAIL Lab, and the Stanford ILIAD Lab for helpful discussions. We thank Google Cloud for providing TPU resources and credits.
This research was partly supported by Volkswagen, an NSF CAREER Award, ARL DCIST CRA W911NF-17-2-0181, DARPA TIAMAT, AFOSR FA9550-22-1-0273, and NSF Grant \#1941722.

%% file: data/7_app.tex
\newpage
\appendix
\onecolumn
\label{app:appendix}
\section{Training Details}

\subsection{Model Architecture}
While our method is applicable to any VLM backbone, we use InternVL2-1B~\cite{chen2024internvl} for both the high-level and low-level policies in our closed-loop experiments. InternVL2-1B is based on Qwen2.5-0.5B-Instruct~\cite{qwen2025qwen25technicalreport} and uses InternViT-300M-448px as its vision encoder. Following the design in~\citep{renz2025simlingovisiononlyclosedloopautonomous}, our low-level policy employs two additional MLP heads for future waypoint prediction. We fine-tune the language model with LoRA and apply full fine-tuning to all remaining parameters.

\subsection{Training Hyperparameters}
Hyperparameters are shown in~\cref{tab:hl_hyperparams}.

\begin{table}[h]
\centering
\begin{minipage}{0.48\textwidth}
\centering
\textbf{High-Level Policy}

\begin{tabular}{lc}
\toprule
\textbf{Hyperparameter} & \textbf{Value} \\
\midrule
Batch Size & $96$ \\
Gradient Accumulation Steps & $2$ \\
Epochs & $20$ \\
Learning Rate & $3\times10^{-5}$ \\
Learning Rate Scheduler & Cosine Decay \\
Betas & $(0.9, 0.999)$ \\
Optimizer & AdamW \\
Warmup steps & 5\% of total steps \\
LoRA alpha & $64$ \\
LoRA r & $32$ \\
LoRA dropout & $0.1$ \\
\bottomrule
\end{tabular}
\end{minipage}
\hfill
\begin{minipage}{0.48\textwidth}
\centering
\textbf{Low-Level Policy}

\begin{tabular}{lc}
\toprule
\textbf{Hyperparameter} & \textbf{Value} \\
\midrule
Batch Size & $120$ \\
Gradient Accumulation Steps & $1$ \\
Epochs & $30$ \\
Learning Rate & $3\times10^{-5}$ \\
Learning Rate Scheduler & Cosine Decay \\
Betas & $(0.9, 0.999)$ \\
Optimizer & AdamW \\
Warmup steps & 5\% of total steps \\
LoRA alpha & $64$ \\
LoRA r & $32$ \\
LoRA dropout & $0.1$ \\
\bottomrule
\end{tabular}
\end{minipage}
\vspace{5pt}
\caption{\textbf{Training Hyperparameters.}}
\label{tab:hl_hyperparams}
\vspace{-5pt}
\end{table}

\subsection{Training and Inference Hardware}
We trained our high-level policy on 8 NVIDIA H100 GPUs for 15 hours and our low-level policy on 4 NVIDIA H200 GPUs for 20 hours. Inference was performed on a single NVIDIA L40 GPU.

\section{Auto-Labeling Pipeline Details}
\subsection{Label Refinement on SimLingo dataset}
\label{app:label_refinement}
In order to imbue language labels from language-annotated datasets, such as the SimLingo dataset, with detailed movement information as described in ~\cref{sec:auto-label}, we provide the vehicle's ego states over a period of three seconds in addition to the original language label in a prompt (see Listing \ref{lst:simlingo_refinement_prompt} for the full prompt) to Gemini 2.5 Flash-Lite.
\definecolor{mytextcolor}{rgb}{0.63,0.08,0.18} 
\definecolor{mybackgroundcolor}{rgb}{0.95, 0.95, 0.95} 
\begin{lstlisting}[
    basicstyle=\ttfamily\footnotesize\color{mytextcolor},
    backgroundcolor=\color{mybackgroundcolor},
    breaklines=true,
    caption=SimLingo Refinement Prompt.,
    label=lst:simlingo_refinement_prompt,
]
You are an expert in vehicle dynamics and driving behavior analysis. Your task is to interpret and refine natural language descriptions of driving behavior by analyzing vehicle ego state data (speed and course over time) to produce a **precise and nuanced behavior summary**. Your output should describe:

1. **Ego State Analysis** - a brief explanation of observed speed and course trends over time.
2. **Refined Driving Behavior Description** - a more specific version of the original description, enhanced with a meaningful modifier _(e.g., **smooth turning**, **wide turn**, **abrupt stop**, **steady lane keeping**)_ and a **driving style**, reflecting the driver's attitude or intent _(e.g., **cautiously**, **normally**, **aggressively**)_

---

## Input Format

**Driving Description:** 
{commentary}

**Ego Vehicle State Sequence** (next 3 seconds from frame {frame_number}):
{ego_states_sequence}

These ego states reflect how the vehicle moved during the described behavior.

> **Note:**  
> - **Course increasing** -> vehicle is adjusting **right**  
> - **Course decreasing** -> vehicle is adjusting **left**

---

## Output Guidelines

Your response should contain two sections:

### 1. Ego State Analysis

Analyze the speed and course sequence:
- Describe speed patterns: Is the vehicle accelerating, decelerating, or maintaining speed?
- Describe course patterns: Is the vehicle turning sharply, smoothly, or going straight?
- Mention time duration and total changes in course or speed.

### 2. Refined Driving Behavior Description

Produce a single, natural-language sentence that:
- Refines the driving description with motion extent (e.g., *smooth*, *sharp*, *wide*, *slight*)
- Adds driving style (e.g., *cautiously*, *normally*, *aggressively*)
- Grounds the refinement in the observable patterns of the ego vehicle states
- Do not change the semantic meaning of the original description. Only use the ego states to refine the description.

---

## Notes

- The refined description must not exceed **20 words**.
- Use **speed trends** to judge acceleration or deceleration patterns.
- Use **course change patterns** to assess turning sharpness or trajectory smoothness.
- If the style cannot be confidently inferred, default to **"normally"**.
- Use **natural, human-readable language**-avoid unnecessary technical jargon.
- The refined description must be a single sentence in present tense and third person (i.e. "The vehicle turns..." or "The car accelerates...")
- If the driving description includes any references to external vehicles, pedestrians or traffic constructs, maintain this information in the final refined description, as well as their distances from the ego vehicle and any descriptiors (i.e. color) if available.
- Do not change the semantic meaning of the original description. Only use the ego states to refine the description.
- If the original description mentions specific maneuvers, i.e. lane changes, retain this information. 
- Unless a turn is explicitly mentioned in the original description, heading changes of 30 degrees or below should be described as **adjustments** to the left or right, and not turns.
\end{lstlisting}

\subsection{NuScenes Meta-Action Labeling}
\label{appendix:meta_action_labels_nuscenes}
Our auto-labeling pipeline is also applicable to real-world datasets without prior language labels. To apply our labeling pipeline to the NuScenes dataset, we begin by splitting trajectories into 2-5 second chunks based on a set of heuristics that define the boundaries of where a specific category of action (e.g., accelerating, changing lanes, or turning) is likely to have occurred. Specifically, we apply a 1D Gaussian blur to the vehicle's speed and course changes over time, and apply splits where the vehicle is stopped, or the vehicle's acceleration or angular velocity are above certain thresholds for an extended period of time.

\begin{figure}[htp]
\centering

\begin{minipage}{0.48\columnwidth}
\centering
\includegraphics[width=0.9\linewidth]{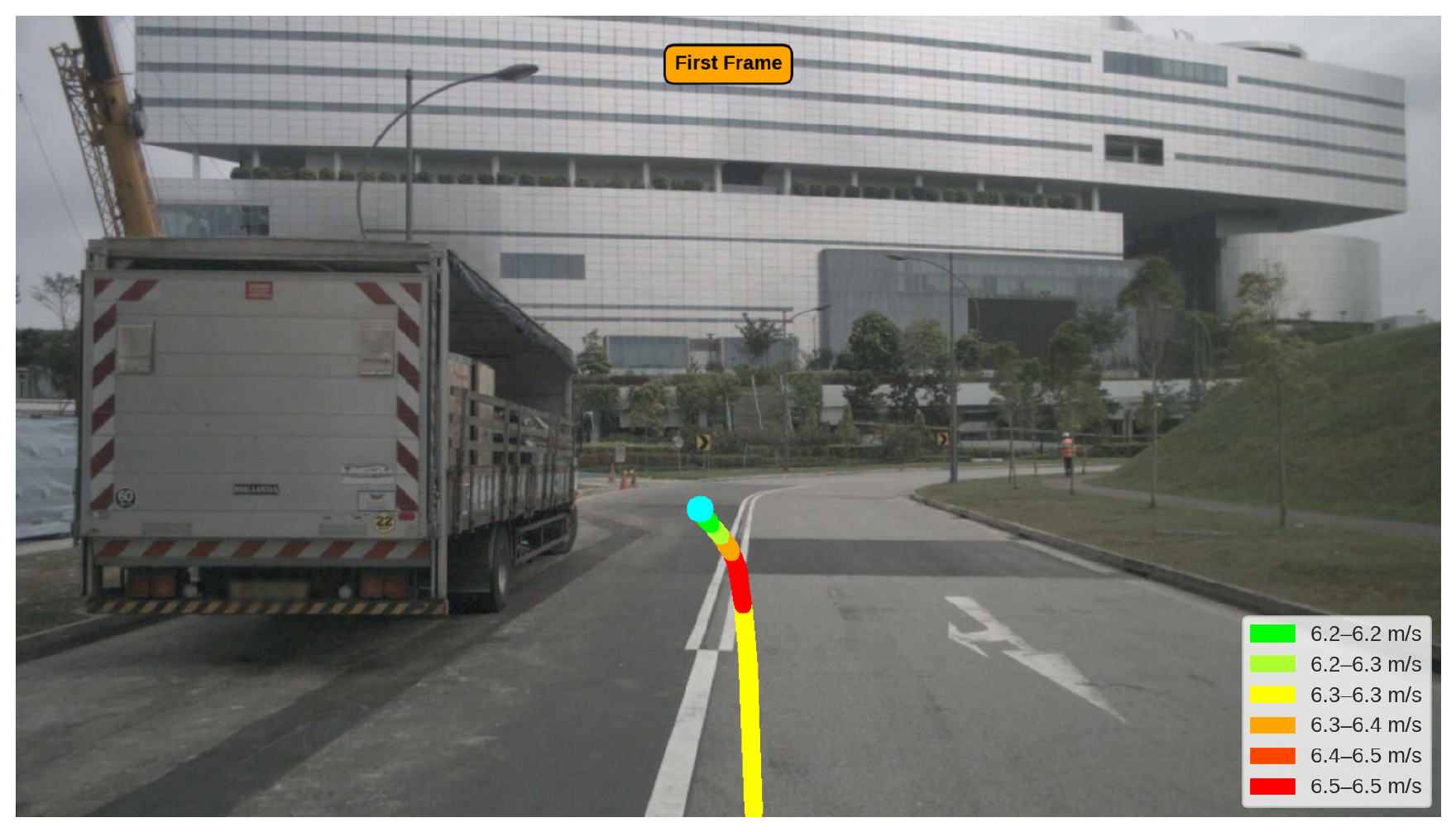}
\includegraphics[width=0.9\linewidth]{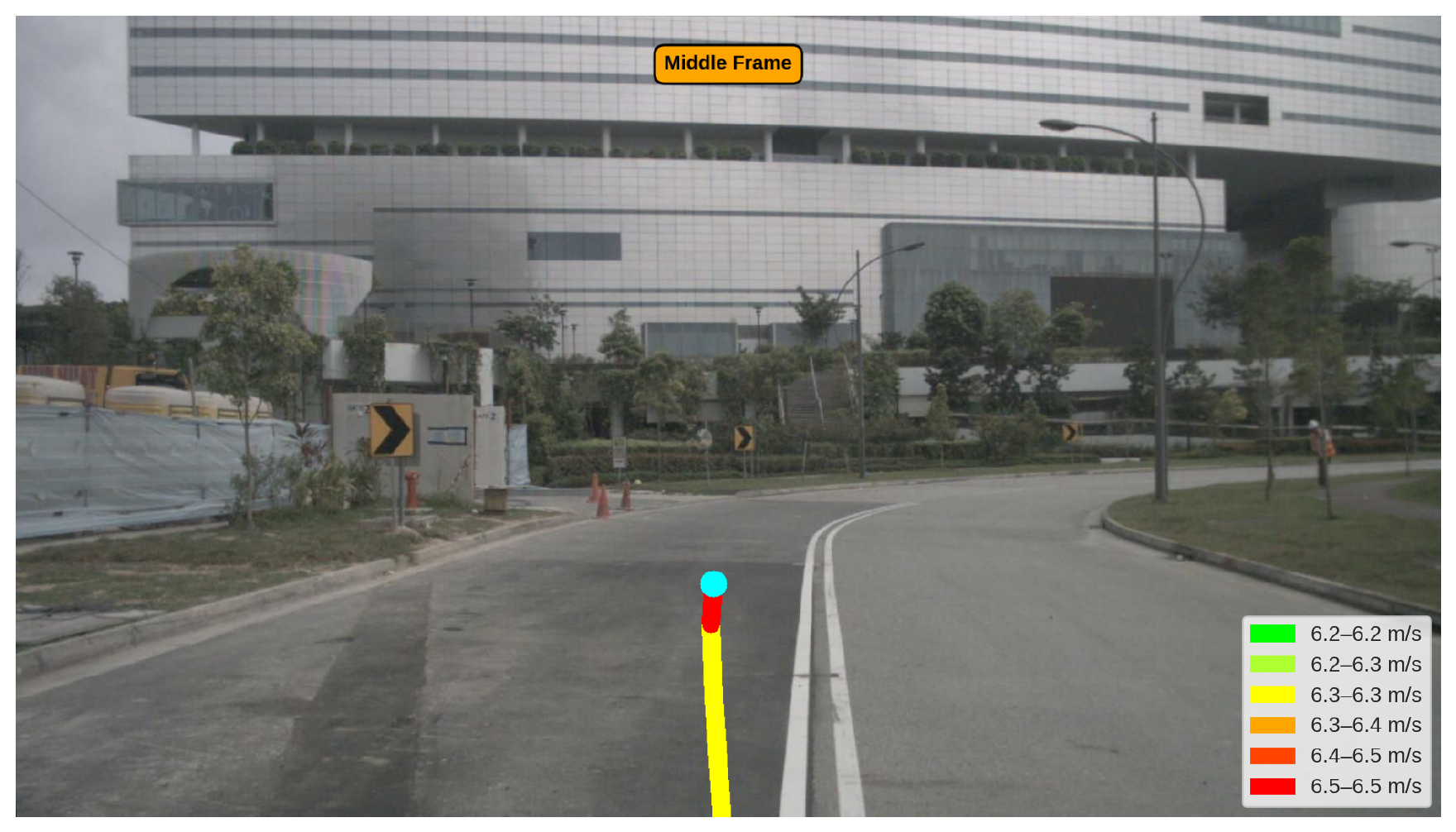}
\caption*{(a) Lane change left example}
\end{minipage}
\hfill
\begin{minipage}{0.48\columnwidth}
\centering
\includegraphics[width=0.9\linewidth]{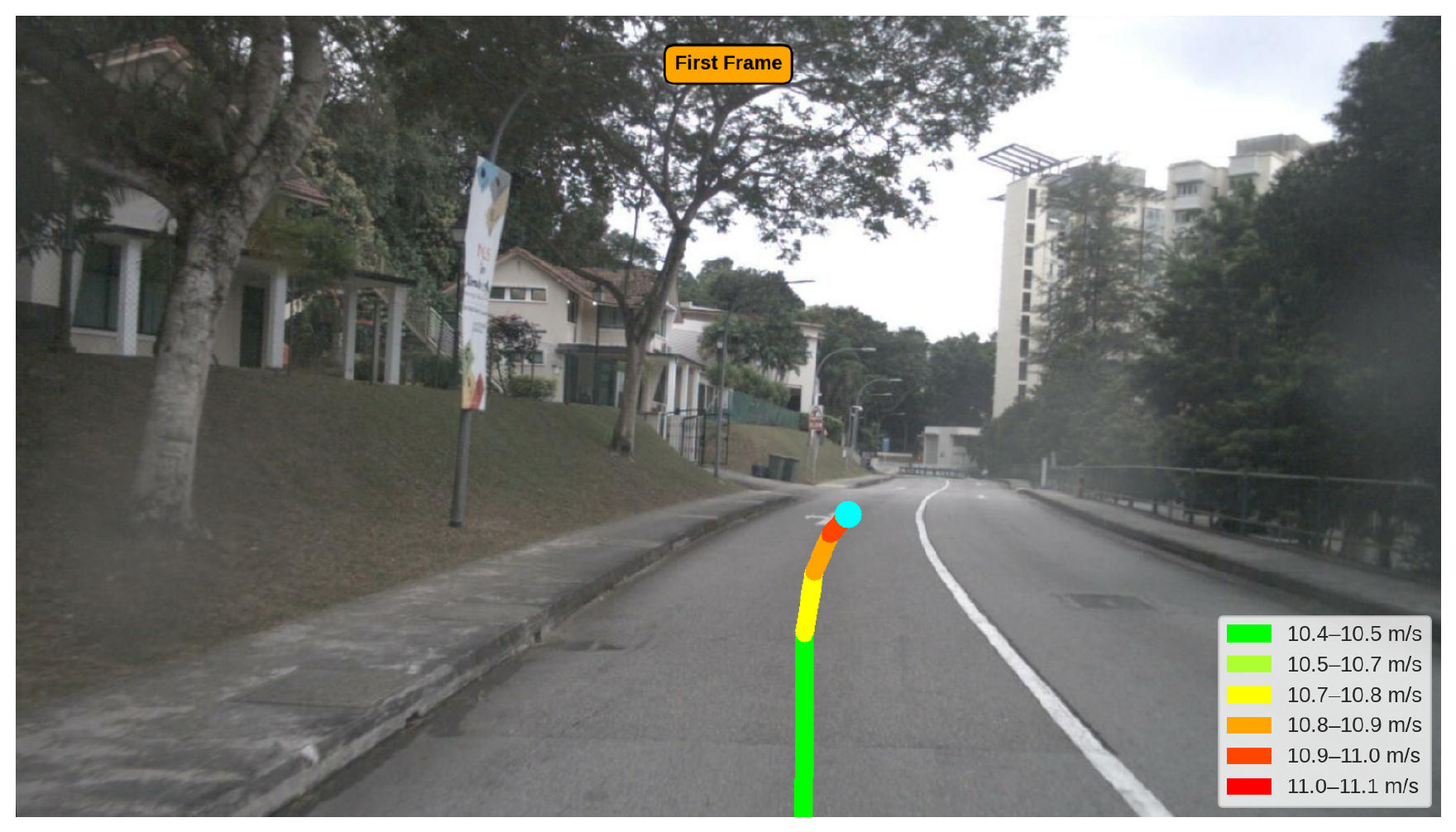}
\includegraphics[width=0.9\linewidth]{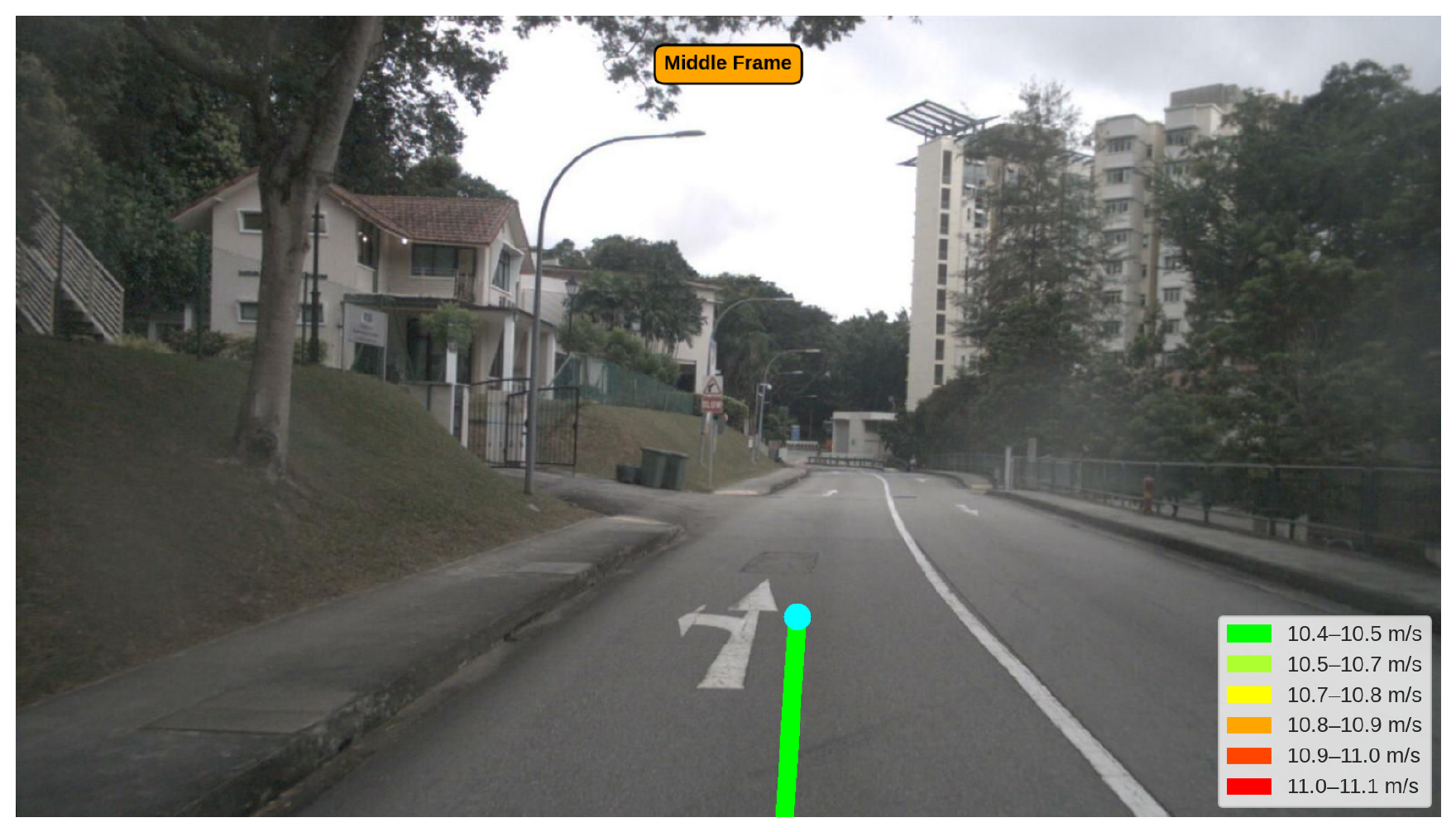}
\caption*{(b) Continue straight example}
\end{minipage}

\caption{\textbf{Input images for meta-action labeling.} The first-round prompt gives a baseline action and the second-round prompt produces a refined meta-action. (a) Example where the baseline action is ``changing lanes left,'' refined to ``The vehicle is smoothly changing lanes left normally.'' (b) Example where the baseline action is ``continuing straight,'' refined to ``The car normally accelerates, then maintains speed while subtly drifting right.''}
\label{fig:nuscenes_label_combined}

\end{figure}

We then utilize the vehicle's camera extrinsic and intrinsic matrices to produce a projection of the vehicle's future trajectory over front camera views from the first and middle frames of the trajectory. These images, the vehicle's ego states and lane IDs over time, and the prompt in Listing \ref{lst:meta_prompt} are provided to Gemini 2.5 Flash-Lite for a baseline categorization stage. We show two examples in~\cref{fig:nuscenes_label_combined}.

\begin{lstlisting}[
    basicstyle=\ttfamily\footnotesize\color{mytextcolor},
    backgroundcolor=\color{mybackgroundcolor},
    breaklines=true,
    caption=Example Meta Action Labeling Prompt.,
    label=lst:meta_prompt,
]
You are an expert in vehicle dynamics and driving behavior analysis. You have been provided two frames from a dashcam video from a vehicle, with a projected green, yellow, and red trajectory overlaid on the first and middle frames of the video of the trajectory that the vehicle is in the process of taking. The images are labelled "First Frame" and "Middle Frame" at the tops of the images. 

Describe:

1. Ego State Analysis:

Analyze the speed and course sequence:
- Describe speed patterns: Is the vehicle accelerating, decelerating, or maintaining speed?
- Describe course patterns: Is the vehicle turning sharply, smoothly, or going straight?
- Mention time duration and total changes in course or speed.

These ego states reflect how the vehicle moved during the described behavior.

> **Note:**  
> - **Course increasing** -> vehicle is moving **right**  
> - **Course decreasing** -> vehicle is moving **left**

{ego_states_text}

2. First frame description: 
- Describe the lane markings in the first frame image, and the projected trajectory's position relative to them at the beginning of the trajectory and at the end. Identify any areas on the road with solid white or yellow lines.
- Are there road markings, signs, or other structures that indicate that the vehicle is at an intersection? 
- Which lane does the trajectory begin in, and which lane does the trajectory end in? 
- Is the red, yellow, and/or green trajectory to the right or left of the lane markings?
- Is the cyan circle to the right or left of the lane markings?
- Is the trajectory curving? If so, which way is the trajectory curving?

3. Middle frame description: 
- Describe the lane markings in the middle frame image, and the projected trajectory's position relative to them at the beginning of the trajectory and at the end. Identify any areas on the road with solid white or yellow lines.
- Are there road markings, signs, or other structures that indicate that the vehicle is at an intersection? 
- Which lane does the trajectory begin in, and which lane does the trajectory end in? 
- Is the red, yellow, and/or green trajectory to the right or left of the lane markings?
- Is the cyan circle to the right or left of the lane markings?
- Is the trajectory curving? If so, which way is the trajectory curving?

4. Consolidated Analysis: 
- Based on your analysis of the first frame image and the middle frame image, which lane does the vehicle begin in, and which lane does it end in? 
- Does this signify a lane change? If so, is the vehicle making a lane change to the left, or a lane change to the right? 
- Alternatively, is the vehicle at an intersection in either frame? Does this signify a turn? Even if the trajectory is curving, consider whether the course change is large enough to be a turn, and whether the vehicle is simply continuing forward to a parallel road.
- If so, is the vehicle turning to the left, or to the right?

5. Vehicle Action: The action that the vehicle is taking. Is the vehicle **turning**, **changing lanes**, or **continuing straight**? If the vehicle is turning or changing lanes, is it doing so to the **left** or to the **right**? Choose from one of the following discrete actions:
- turning left
- turning right
- changing lanes left
- changing lanes right
- continuing straight
- completely stationary
- making a U-Turn

Notes:
- The cyan circle denotes the **end** of the trajectory.
- The trajectory begins at the **bottom** of the image.
- A turn is defined as a full turn at an intersection.
- Otherwise, if the trajectory is simply following a curve in the road, describe this as **continuing straight**
- If the trajectory is **continuing straight** through an intersection, describe this as **continuing straight**
- If the vehicle has crossed a lane marking, it is most likely making a lane change.
- There may be no visible trajectory projected, in which case the vehicle is most likely moving very slowly or stationary.
- Identify only the lane markings that are clearly discernible.
- Small course changes of fewer than 4 degrees most likely indicate that the vehicle is **continuing forward**.
- Large course changes over 50 degrees likely indicate that the vehicle is **turning**. 
- Small velocities below 1.0 meters per second likely indicate that the vehicle is stationary.

Lane information: {lane_information}
\end{lstlisting}
We then provide the output of Listing \ref{lst:meta_prompt}, the vehicle's ego states, and the prompt in Listing \ref{lst:meta_stage2_prompt} to Gemini 2.5 Flash-Lite in a refinement step that imbues the resulting meta-action with more detailed movement information.
\begin{lstlisting}[
    basicstyle=\ttfamily\footnotesize\color{mytextcolor},
    backgroundcolor=\color{mybackgroundcolor},
    breaklines=true,
    caption=Example Meta Action Labeling Prompt.,
    label=lst:meta_stage2_prompt,
]
# Driving Behavior Refinement Prompt

You are an expert in vehicle dynamics and driving behavior analysis. Your task is to interpret and refine natural language descriptions of driving behavior by analyzing vehicle ego state data (speed and course over time) to produce a **precise and nuanced behavior summary**. Your output should describe:

1. **Ego State Analysis** -> a brief explanation of observed speed and course trends over time.
2. **Refined Driving Behavior Description** - a more specific version of the original description, enhanced with a meaningful modifier _(e.g., **smooth turning**, **wide turn**, **abrupt stop**, **steady lane keeping**)_ and a **driving style**, reflecting the driver's attitude or intent _(e.g., **cautiously**, **normally**, **aggressively**)_

---

## Input Format

**Driving Description:**  
{driving_description}

**Ego Vehicle States:**  
{ego_state_sequence}

These ego states reflect how the vehicle moved during the described behavior.

> **Note:**  
> - **Course increasing** -> vehicle is moving **right**  
> - **Course decreasing** -> vehicle is moving **left**

---

## Output Guidelines

Your response should contain two sections:

### 1. Ego State Analysis

Analyze the speed and course sequence:
- Describe speed patterns: Is the vehicle accelerating, decelerating, or maintaining speed?
- Describe course patterns: Is the vehicle turning sharply, smoothly, or going straight?
- Mention time duration and total changes in course or speed.

### 2. Refined Driving Behavior Description

Produce a single, natural-language sentence that:
- Refines the driving description with motion extent (e.g., *smooth*, *sharp*, *wide*, *slight*)
- Adds driving style (e.g., *cautiously*, *normally*, *aggressively*)
- Grounding the refinement in the observable patterns of the ego vehicle states

---

## Notes

- The refined description must not exceed **20 words**.
- Use **speed trends** to judge acceleration or deceleration patterns.
- Use **course change patterns** to assess turning sharpness or trajectory smoothness.
- If the style cannot be confidently inferred, default to **"normally"**.
- Use **natural, human-readable language**-avoid unnecessary technical jargon.
- If the driving description is "The vehicle is continuing straight", describe any left or right movements as "adjusting left" or "adjusting right" respectively. Do not describe this as turning.
\end{lstlisting}

\subsection{NuScenes Reasoning Labeling}
\label{appendix:reasoning_nuscenes}
To produce reasoning traces for the NuScenes dataset, we provide a front camera view, as well as the prompt in Listing \ref{lst:reason_prompt} to Gemini 2.5 Flash-Lite. 
\begin{lstlisting}[
    basicstyle=\ttfamily\footnotesize\color{mytextcolor},
    backgroundcolor=\color{mybackgroundcolor},
    breaklines=true,
    caption=Example Reasoning Labeling Prompt.,
    label=lst:reason_prompt,
]
You are an expert in autonomous driving planning. Given a first person dashcam view from a car and the car's future action, describe the following:

Future action: {meta_action}

1. Provide a sentence of justification for the car's future action.
- A concrete example is as follows: There is a car in the oncoming lane and an accident ahead of me, so I should wait within my lane until the oncoming car is clear.

2. Behavior description of critical objects: describe the current status and intent for the 2-3 most important critical objects in the image (e.g. pedestrians, vehicles, cyclists, stop signs, traffic lights, construction cones, etc.) in 3 sentences or fewer. 
- A concrete example is as follows: The pedestrian is currently standing on the sidewalk, looking toward the road, and maybe preparing to cross the street. The vehicle is currently ahead of me, moving in the same direction, and its future trajectory suggests it will continue straight.
\end{lstlisting}

\subsection{VLM Zero-shot Prompt}
\label{appendix:zero_shot_gemini}
To compare the zero-shot capabilities of an off-the-shelf VLM on our meta-action prediction task, we provide the prompt in Listing \ref{lst:zero_shot_prompt} to Gemini 2.5 Flash-Lite (the same model used for labeling), together with the vehicle's visual observation. 
\begin{lstlisting}[
    basicstyle=\ttfamily\footnotesize\color{mytextcolor},
    backgroundcolor=\color{mybackgroundcolor},
    breaklines=true,
    caption=Zero-shot Prompt Provided to Gemini 2.5 Flash-Lite.,
    label=lst:zero_shot_prompt,
]
You are an autonomous driving assistant. Your task is to generate a driving behavior plan based on:
A front-view camera image
A sequence of historical ego states taken at 0.5 Hz over the past 6 seconds
The current speed of the vehicle
A routing command.

Inputs:
Image: <first person image from a dashcam view>
Speed history: 0.0 m/s 0.0 m/s 0.0 m/s 
Heading history: 165.8 degrees 165.8 degrees 165.8 degrees
Current speed: 0.0 m/s
Routing command: Follow the road.

Output:

1. Behavior description of critical objects: describe the current movement and appearance of all external agents in the scene, as well as their positioning relative to the ego vehicle. 
Example Output:
"Red car, in one lane to the left, traveling same direction, at 6.1 m/s. Female pedestrian, in crosswalk, travelling opposite direction, at 2.1 m/s."

2. Driving Behavior Plan:
Produce a driving behavior plan (no more than 20 words) that includes:
Speed behavior - Will the vehicle accelerate, maintain speed, or decelerate?
Heading behavior - Describe the expected heading change (e.g., continue straight, turn slightly right, make a sharp left).
Driving style - Reflect the style (e.g., cautiously, smoothly, assertively).
Respond with a single natural language sentence summarizing the driving behavior.
Example Output:
"The car decelerates smoothly and makes a slight right turn, driving normally to follow the blue SUV."

Notes:
- The driving behavior plan must be in present tense and third person (i.e. "The vehicle...")
\end{lstlisting}

\section{Experiment Details}

\subsection{Full description of baselines}
\label{app:baselines}

\textbf{SimLingo~\citep{renz2025simlingovisiononlyclosedloopautonomous}.} A vision-only VLM framework that addresses closed-loop driving, vision-language understanding, and language-action alignment, relying solely on cameras and avoiding costly sensors such as LiDAR. SimLingo additionally leverages ``action-dreaming'' data, which is counterfactual data used to improve its language following capabilities. SimLingo is currently the top method on the CARLA 2.0 leaderboard.

\textbf{DriveMoE~\citep{yang2025drivemoemixtureofexpertsvisionlanguageactionmodel}.} Built upon the $\pi_0$ foundation model~\citep{black2024pi0visionlanguageactionflowmodel}, DriveMoE employs a mixture-of-experts architecture with a scene-specialized vision MoE and a skill-specialized action MoE to achieve adaptive decision making for autonomous driving. 
\textbf{ORION~\citep{fu2025orion}.} A holistic E2E framework that integrates a QT-Former for long-term history aggregation, an LLM for driving scenario reasoning, and a generative planner for precise trajectory prediction. ORION further aligns reasoning and action spaces, enabling unified optimization across both planning and visual question answering, though at the cost of greater complexity and computational demand.

\textbf{AutoVLA~\citep{zhou2025autovla}.} A method that enhances a pretrained VLM with a physical action codebook for vehicle motion, effectively bridging semantic reasoning and low-level control. 

\textbf{PARA-Drive~\citep{weng2024paradrive}.} A modular end-to-end autonomous driving model that uses bird's-eye-view features and is parallelized to improve runtime efficiency, which offers a comparison alternative modular architecture to our hierarchical structure for comparison. 

\textbf{TOKEN~\citep{tokenizetheworld}.} A method that tokenizes sensory inputs into object-centric tokens using an end-to-end driving model, PARA-Drive, trained with various driving tasks to enforce good representations. This method leverages explicit structure inspired by traditional driving stacks rather than leveraging VLM priors to make good driving decisions.

\textbf{DiMA-VAD~\citep{hegde2025distillingmultimodallargelanguage}.} DiMA-VAD distills knowledge from a VLM into a driving model through jointly training the VLM and a vision-based planner on a set of surrogate driving understanding and prediction tasks rather than directly using the VLM as a base model. 

\textbf{Agent-Driver~\citep{mao2023language}.} This method decomposes the LLM's tasks into using a tool library to process sensory inputs, structuring information in a memory module, and performing motion planning with a reasoning engine. This method explicitly performs many of the reasoning steps that are implicitly included in our auto-labeling pipeline.

\begin{table*}[h!]
\centering
\setlength{\tabcolsep}{5pt}

\resizebox{\textwidth}{!}{
\begin{tabular}{lcccccccc|cc}
\toprule
\multirow{3}{*}{\textbf{Method}} & \multirow{3}{*}{\textbf{Sensors}} & \multirow{3}{*}{\textbf{DS} $\uparrow$} & \multirow{3}{*}{\textbf{SR(\%)} $\uparrow$} & \multicolumn{6}{c}{\textbf{Ability}$\uparrow$} \\[0.5ex]
\cline{5-10}
& & & & \rule{0pt}{3.5ex} Merging & \makecell{Over-\\taking} & \makecell{Emergency\\Brake} & \makecell{Give\\Way} & \makecell{Traffic\\Sign} & Mean \\
\midrule
DriveMoE & M & 74.22 & 48.64 & 34.67 & 40.00 & 65.45 & 40.00 & 59.44 & 47.91\\
ORION & M & 77.74 & 54.62 & 25.00 & 71.11 & 78.33 & 30.00 & 69.15 & 54.72\\
AutoVLA & M & 78.84 & 57.73 & - & - & - & - & - & -\\
SimLingo & S & {85.94} & {66.82} & \textbf{57.50} & {60.00} & {76.67} & 50.00 & {73.16} & {63.46}\\
\hline
\rowcolor{blue!10}
\MethodName{} (Ours) & S & {\textbf{90.71}} & {\textbf{73.64}} & {56.25} & {\textbf{84.44}} & {\textbf{81.67}} & \textbf{60.00} & {\textbf{81.05}} & {\textbf{72.68}} \\
\bottomrule
\end{tabular}
}
\caption{\textbf{Evaluation of \MethodName{} on Bench2Drive.} Metrics include Driving Score (DS), Success Rate (SR\%), and specialized abilities (Merging, Overtaking, Emergency Brake, Give Way, Traffic Sign) with overall Mean performance. Compared to the state-of-the-art, \MethodName{} outperforms the best performing baseline (SimLingo). M/S refers to Multi-camera/Single camera.}
\vspace{-10pt}
\label{tab:bench2drive}
\end{table*}

\begin{table}[t]
    \centering
    \small
    \begin{tabular}{lccc}
    \toprule
         & & \multicolumn{2}{c}{\textbf{Driving score} $\uparrow$} \\
\cline{3-4}
\textbf{Long-tail Scenario} & \# Routes & Simlingo & SteerVLA \\
        \midrule
        Illegally Parked Vehicle & 10 & 96.00 & \textbf{100.00} \\
        Adjacent Door Opening    & 5 & \textbf{92.00} & \textbf{92.00} \\
        Roadside Cyclist         & 10 & \textbf{88.00} & \textbf{88.00} \\
        Construction Zone        & 10 & 62.66 & \textbf{96.50} \\
        Traffic Accident         & 10 & 68.91 & \textbf{86.33} \\
        Jaywalking Pedestrian    & 10 & \textbf{100.00} & 95.00 \\
        Crossing Vehicle Runs Red & 5 & 71.94 & \textbf{72.06} \\
        Control Loss & 5 & \textbf{100.00} & \textbf{100.00} \\
        Hard Brake & 5 & \textbf{100.00} & 98.86 \\
        Blocked Intersection & 5 & 76.86 & \textbf{100.00} \\
        Yield to Emergency Vehicle & 5 & 70.00 & \textbf{76.00} \\
        \midrule
        \rowcolor{blue!10}
        \textbf{Mean (route-weighted)} & & 83.87 & \textbf{91.91} \\
        \bottomrule
    \end{tabular}
    \vspace{5pt}
    \caption{\textbf{Performance comparison across long-tail driving scenarios on Bench2Drive-LongTail.} \MethodName{} demonstrates strong long-tail performance across various scenarios in the }
    \label{tab:long_tail_comparison}
    \vspace{-8pt}
\end{table}

\subsection{Bench2Drive-LongTail}
\label{app:long-tail}
To evaluate the long-tail performance of \MethodName{}, we introduce a long-tail subset of Bench2Drive. We focus on 11 categories: 
\vspace{-10pt}
\begin{enumerate}
    \setlength{\parskip}{0cm}
    \setlength{\itemsep}{0cm}
    \item \textbf{Crossing vehicle runs red. } A vehicle moving perpendicular to the ego-vehicle runs a red light. The ego-vehicle must recognize that it should wait for the vehicle before it proceeds. 
    \item \textbf{Yield to emergency vehicle. } An emergency vehicle is driving down the street. The ego-vehicle must yield and wait for the emergency vehicle to pass. 
    \item \textbf{Traffic accident. } A traffic accident has occurred, and the ego-vehicle must avoid the scene while interacting safely with other vehicles. 
    \item \textbf{Roadside cyclist. } A cyclist it traveling along the same road as the ego-vehicle. The ego-vehicle must safely avoid the cyclist. 
    \item \textbf{Adjacent door opening. } A vehicle on the side of the road opens its door into traffic. The ego-vehicle must safely navigate out of the situation while interacting safely with other agents. 
    \item \textbf{Jaywalking pedestrian. } A pedestrian crosses the street at a non-designated crossing point. The ego-vehicle must slow to wait for the pedestrian to pass. 
    \item \textbf{Construction zone.} A construction zone has modified the flow of traffic. The ego-vehicle must avoid the construction zone and merge back into the normal traffic flow. 
    \item \textbf{Hard brake.} A sudden obstacle in the road causes the vehicle to need to brake abruptly. 
    \item \textbf{Illegally parked vehicle. } A vehicle is parked illegally, obstructing the roadway. The ego-vehicle must reason that the vehicle is in fact parked and navigate around it. 
    \item \textbf{Control loss. } The ego-vehicle encounters an area of poor traction and loses control. It must recover control without collision. 
    \item \textbf{Blocked intersection. } Traffic blocks an intersection. The ego-vehicle must reason about the best course of action.
\end{enumerate}

\begin{table}
\centering
\small
\renewcommand{\arraystretch}{1.6}
\setlength{\tabcolsep}{5pt}
\begin{tabular}{lccc|c}
\toprule
\multirow{2}{*}{\textbf{Method}} & \multicolumn{4}{c}{\textbf{Traj L2 (m)} $\downarrow$} \\
\cline{2-5}
& 1s & 2s & 3s & Mean \\
\hline
\makecell{TOKEN \\ \citep{tokenizetheworld}} & 0.26 & 0.70 & 1.46 & 0.81 \\
\makecell{PARA-Drive \\ \citep{weng2024paradrive}} & 0.26 & 0.59 & 1.12 & 0.66  \\
\makecell{DiMA-VAD \\ \citep{hegde2025distillingmultimodallargelanguage}} &  0.18 & 0.48 & 1.01 & 0.56 \\
\makecell{GPT-Driver \\ \citep{mao2023gpt}} & 0.20 & 0.40 & 0.70 & 0.44 \\
\makecell{Agent-Driver \\ \citep{mao2023language}} & \textbf{0.16} & \textbf{0.34} & \textbf{0.61} & \textbf{0.37} \\
\hline
\rowcolor{blue!10}
\makecell{SteerVLA (Ours)} & 0.18 & 0.39 & 0.63 & 0.40 \\
\bottomrule
\end{tabular}
\caption{\textbf{Open-loop comparison of \MethodName{} on the NuScenes planning benchmark.} \MethodName{} achieves comparable L2 error compared to state-of-the-art methods.}
\vspace{-10pt}
\label{tab:nuscenesplanning}
\end{table}

The full list of driving scenarios in Bench2Drive is available here. We select what we believe to be the long-tail subset from the list~\citep{jia2024bench2drivemultiabilitybenchmarkingclosedloop}. 

We compute the route-weighted mean as $Mean = \frac{\sum((\# \text{routes per category}) \cdot (\text{DS per category})}{\text{\# routes total}}$, where DS is driving score. 

\subsection{Raw values of Performance on Bench2Drive}
\label{app:raw_values}
In addition to~\cref{fig:bench2drive} and~\cref{fig:longtail_bench2drive}, we provide the raw scores of the baselines and \MethodName{} evaluated on Bench2Drive and Bench2Drive-LongTail, provided in \cref{tab:bench2drive} and~\cref{tab:long_tail_comparison}. 

\subsection{Failure Cases in Closed-Loop Evaluation}
\label{app:failure_cases}
However, we still observe failure cases for \MethodName{}, which primarily fail into two categories. First, some failures arise from the use of a single front-view camera, which limits visibility of vehicles approaching from the sides or rear (e.g. when yielding to an emergency vehicle). Incorporating multi-view camera inputs is a promising direction for future work. Second, \MethodName{} exhibits limited recovery behavior once it enters out-of-distribution states following an incorrect action. For example, when an early or aggressive lane change places the vehicle in an unexpected position relative to surrounding traffic, the policy may fail to recover safely. This limitation is likely due to insufficient coverage of non-optimal behaviors in the training data, which predominantly consists of expert demonstrations. In future work, we plan to address this issue by incorporating co-training or additional supervision from real-world data, where state distributions are more diverse and include recovery behaviors.

\subsection{Open-Loop Evaluation on Real-World Data}
\label{app:nuscenes}
We additionally evaluate \MethodName{} in an open-loop setting on the NuScenes planning benchmark~\citep{nuscenes} to assess performance on real-world driving data. We adopt stronger VLM backbones than in our simulation experiments: Gemma-3 4B~\citep{gemmateam2025gemma3technicalreport} as the high-level policy and PaliGemma~\citep{beyer2024paligemmaversatile3bvlm} as the low-level policy. For the low-level policy, we follow the approach of \citet{kimOpenVLAOpenSourceVisionLanguageAction2024}, repurposing rarely used tokens to represent discretized actions, where each dimension is divided into 512 uniform bins over the normalized range $[-1, 1]$ based on dataset statistics~\citep{teamOctoOpenSourceGeneralist2024, brohanRT2VisionLanguageActionModels2023}. Since NuScenes does not provide language annotations, we apply our automatic labeling pipeline to generate grounded meta-actions and reasoning traces directly from raw driving data, which are used to supervise both the high-level and low-level policies. Details on data labeling are provided in \cref{appendix:meta_action_labels_nuscenes,appendix:reasoning_nuscenes}. All models are trained and evaluated on the official NuScenes training and validation splits.

We apply the full auto-labeling pipeline to the NuScenes dataset and evaluate \MethodName{} against several baselines on the NuScenes planning benchmark (see~\cref{tab:nuscenesplanning}). The policy is executed at 2 Hz, and performance is measured using L2 trajectory error over prediction horizons of 1, 2, and 3 seconds. We compare \MethodName{} with PARA-Drive~\citep{weng2024paradrive}, TOKEN~\citep{tokenizetheworld}, DiMA-VAD~\citep{hegde2025distillingmultimodallargelanguage}, and Agent-Driver~\citep{mao2023language}, which represent a range of modular, token-based, distilled, and LLM-guided driving approaches.

As shown in~\cref{tab:nuscenesplanning}, \MethodName{} achieves performance comparable to or better than existing methods across all horizons, indicating that our framework generalizes effectively to real-world driving data despite being primarily evaluated in closed-loop simulation.

\renewcommand{\thesubsection}{A\arabic{subsection}}